\documentclass[letterpaper, 10 pt, conference]{ieeeconf}  

\IEEEoverridecommandlockouts                              

\usepackage{amsmath}
\usepackage{amssymb}
\usepackage{graphicx}
\usepackage{booktabs}
\usepackage{multirow}
\usepackage{colortbl}
\usepackage{pifont}
\usepackage{xcolor}
\usepackage[
  colorlinks=true,
  linkcolor=cyan,
  urlcolor=cyan,
  citecolor=cyan
]{hyperref}
\usepackage{algorithm}
\usepackage{algorithmic}

\title{\LARGE \bf
GeoTeacher: Geometry-Guided Semi-Supervised 3D Object Detection
}

\author{Jingyu Li$^{1,2*}$, Xiaolong Zhao$^{3*}$, Zhe Liu$^{4}$, Wenxiao Wu$^{5,2}$, Li Zhang$^{1,2\dagger}$%
\thanks{$\dagger$ Corresponding author}%
\thanks{$^{1}$ The authors are with Fudan University.}%
\thanks{$^{2}$ The authors are with Shanghai Innovation Institute.}%
\thanks{$^{3}$ The author is with Tongji University.}%
\thanks{$^{4}$ The author is with Hong Kong University.}%
\thanks{$^{5}$ The author is with Huazhong University of Science and technology.}
}
\begin{document}

\maketitle
\thispagestyle{empty}
\pagestyle{empty}

\begin{abstract}
Semi-supervised 3D object detection~(SS3D), aiming to explore unlabeled data for boosting 3D object detectors, has emerged as an active research area in recent years. 
Some previous methods have shown substantial improvements by either employing heterogeneous teacher models to provide high-quality pseudo labels or enforcing feature-perspective consistency between the teacher and student networks.
However, these methods overlook the fact that the model usually tends to exhibit low sensitivity to object geometries with limited labeled data, making it difficult to capture geometric information, which is crucial for enhancing the student model’s ability in object perception and localization.
In this paper, we propose GeoTeacher to enhance the student model's ability to capture geometric relations of objects with limited training data, especially unlabeled data. 
We design a keypoint-based geometric relation supervision module that transfers the teacher model’s knowledge of object geometry to the student, thereby improving the student’s capability in understanding geometric relations.
Furthermore, we introduce a voxel-wise data augmentation strategy that increases the diversity of object geometries, thereby further improving the student model’s ability to comprehend geometric structures. 
To preserve the integrity of distant objects during augmentation, we incorporate a distance-decay mechanism into this strategy.
Moreover, GeoTeacher can be combined with different SS3D methods to further improve their performance.
Extensive experiments on the ONCE and Waymo datasets indicate the effectiveness and generalization of our method and we achieve the new state-of-the-art results. 
Code will be available at \url{https://github.com/LogosRoboticsGroup/GeoTeacher}
\end{abstract}

\section{Introduction}

3D object detection, which detects objects of interest within 3D scenes, is essential for autonomous driving, robotics, and various other applications. However, achieving high-performance detection typically depends on large and fully labeled datasets, which are quite costly and time-intensive. In order to solve this problem, semi-supervised 3D object detection~(SS3D) is born on demand, which intends to train object detectors based on labeled data as well as easy-to-obtain unlabeled data. 

Current SS3D methods generally follow the 2D approach, utilizing a teacher-student framework where the teacher model generates pseudo-labels to train the student model. Some prior methods improve the student model’s performance by incorporating additional information or designing sophisticated networks for the teacher model to generate high-quality pseudo labels.
For instance, A-Teacher~\cite{a-teacher} and X-ray Teacher~\cite{gambashidze2024weak} utilize multi-frame fusion to obtain pseudo labels, while ProficientTeacher~\cite{yin2022semi} introduces test-time augmentation and a refinement network.
Recently, some methods have realized the importance of enforcing feature-level supervision to enhance student model performance. For example, NoiseDet~\cite{Chen_2023_ICCV} introduces a dense Gaussian feature constraint to improve the transferability of learned features, and DPKE~\cite{han2024dual} enforces feature consistency on downsampled points for indoor scenarios. 
However, the aforementioned methods neglect the exploration and integration of higher-order information within the object's internal geometry, which is particularly crucial for enhancing the student model's perceptual ability when only limited labeled data is available. This limitation hinders their potential for achieving higher performance.

To this end, we propose a novel semi-supervised 3D object detection method named GeoTeacher, which can be easily combined with existing semi-supervised approaches. GeoTeacher aims to guide the student model learning the inherent information contained within object geometries, particularly geometric information, derived from unlabeled data.
Specifically, we design a keypoint-based geometric relation supervision module. We select keypoints~(e.g., center points and corner points) that reflect the geometric characteristics of objects and treat the correlations among them as geometric properties. Unlike pseudo labels or features, the geometric relations could effectively represent the unique structure of each object. To mitigate the influence of unreliable knowledge, we employ pseudo-label scores to weigh the geometric knowledge of different objects accordingly.

In addition, we introduce a voxel-wise augmentation data strategy to increase the geometric diversity of objects, applicable to both labeled and unlabeled data.
Different from prior approaches~\cite{liu2023hierarchical,Wu_Peng_Xie_Hou_Lin_Huang_Liu_Cai_Ouyang_2024} that partition the entire point cloud scene, our method focuses on decomposing individual objects into smaller constituent parts. By doing this, we are able to modify their geometry flexibly while ensuring they remain detectable. The student model can achieve more accurate object detection when incorporated with our geometric relation supervision.
Since distant objects are often represented by sparse point clouds and are more challenging to detect, we introduce a distance-decay strategy to preserve their geometry during augmentation.

The main contributions of this paper can be summarized as follows: 
{\bf (i)}~We present a novel SS3D method, named \textit{GeoTeacher}. GeoTeacher guides the student model to learn geometric information from both data and supervision perspectives. 
{\bf (ii)}~We design a geometric relation supervision model and a distant-decay voxel-wise data augmentation strategy to facilitate the student's ability to learn geometric knowledge.
{\bf (iii)}~Extensive experiments demonstrate that our method can effectively combine with existing SS3D approaches and achieves performance on par with or surpassing the current state-of-the-art semi-supervised 3D object detection methods. 

\section{Related work}
\subsection{LiDAR-based 3D object detection}
3D object detection~\cite{huang2020epnet,liu2022epnet++,li2023dds3d,qi2017pointnet,shi2023pv,park2022detmatch,wang2023dsvt,qin2023supfusion} has made rapid progress in the past few years and is widely used in the field of autonomous driving and navigation robot. In terms of point cloud representation: point-based~\cite{zhang2022not,shi2019pointrcnn,zheng2021se} and voxel-based~\cite{liu2020tanet,zhou2018voxelnet,yan2018second,liu2024lion,liu2025unilion}. Point-based methods directly use irregular point cloud to extract features. 
Voxel-based methods adopt voxelization to make point clouds regular, making it possible to extract features using asymmetric functions, e.g., convolution. 
Besides this, PV-RCNN~\cite{Shi_2020_CVPR} achieves higher performance by combining multi-scale voxel and point cloud features.
DetZero~\cite{ma2023detzero} proposes a combination of an offline tracker and a multi-frame detector, achieving significant performance improvements.
However, 3D object detection depends on large-scale labeled data, which is costly to obtain. Our method aims to leverage unlabeled data to enhance detection performance.

\subsection{Semi-supervised learning}
Compared to supervised learning, SSL is designed to train models with a small amount of labeled data and abundant unlabeled data, increasing the difficulty of the task. Previous works~\cite{zhang2021flexmatch,wang2022freematch,NEURIPS2019_mixmatch,NEURIPS2020_06964dce} are mainly divided into two categories: consistency regularization and pseudo-labeling. 
Temporal Ensembling~\cite{laine2016temporal} first proposes consistency regularization, which aims to minimize the difference between predictions from different steps. 
Mean Teacher~\cite{tarvainen2017mean} contains two branches, i.e., teacher and student, with the same architecture. The teacher model is an EMA of the student model.  
Pseudo-labeling methods~\cite{li2020dividemix,arazo2020pseudo} usually first train the model with labeled data and then iteratively generate pseudo labels to unlabeled data through self-training to improve performance. 
Noisy Student~\cite{Xie_2020_CVPR} injects noise into the student model, which is then supervised by the teacher model. In order to obtain more accurate pseudo labels.
\begin{figure*}[t]
    \centering
    \includegraphics[width=\linewidth]{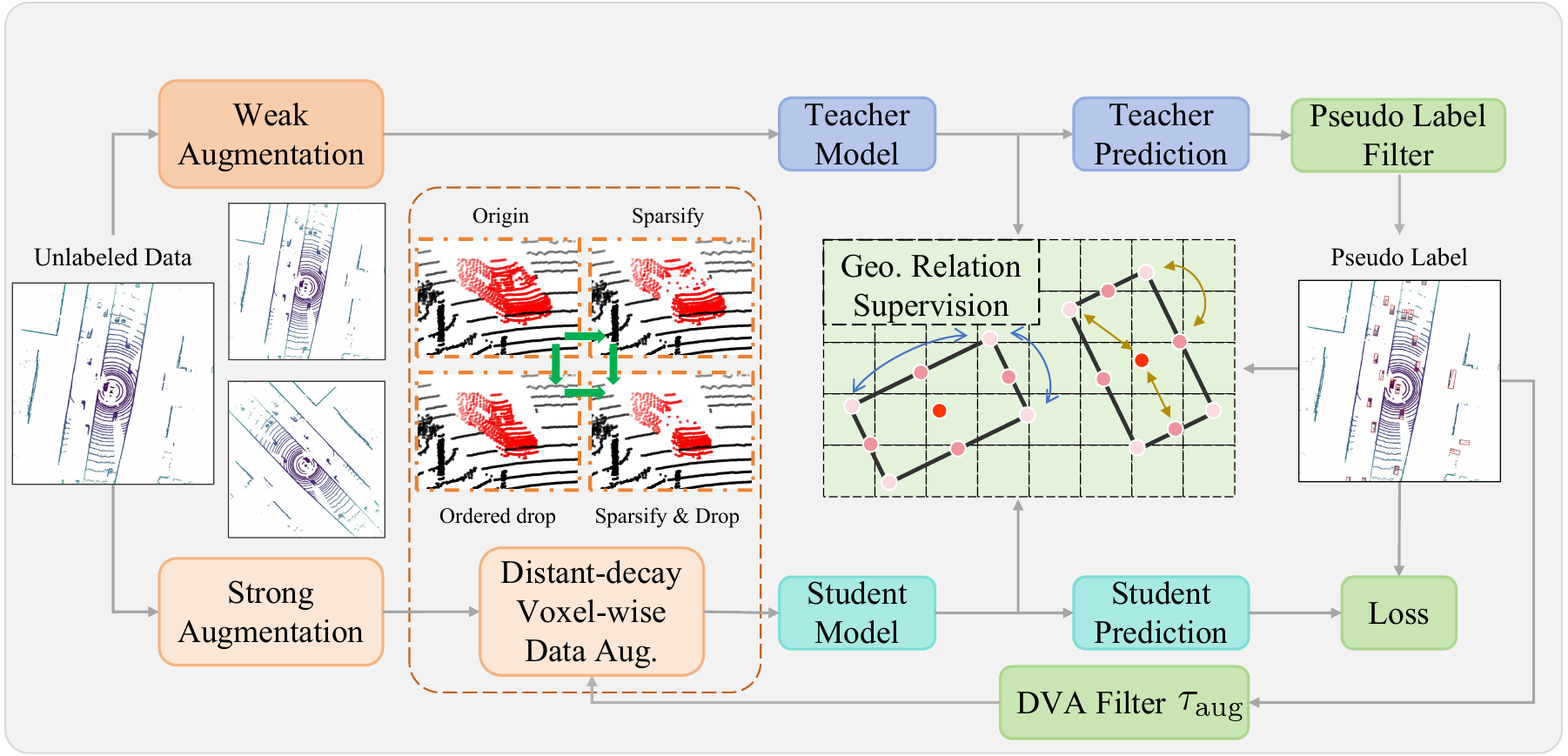}
    \caption{\textbf{The framework of our GeoTeacher.} After applying strong augmentation, we further introduce our proposed data augmentation strategy to enhance the geometric diversity of objects. The student model is supervised with both the standard semi-supervised loss and our geometric relation supervision loss, which effectively improves the student model’s ability to comprehend the intrinsic geometric information of objects.}
    \label{fig:pipeline}
\end{figure*}

\subsection{Semi-supervised 3D object detection}
Most traditional SSL techniques are proposed for classification tasks, but also benefit semi-supervised object detection~\cite{li2022pseco,zhu2024doubly,zhou2022dense,xu2021end,li2023dds3d}, especially SS3D~\cite{zhan2024csot}.
SESS~\cite{zhao2020sess} and 3DIoUMatch~\cite{wang20213dioumatch} are two pioneers in exploring SS3D task. SESS uses Mean Teacher paradigm in point-based 3D object detection, introducing asymmetric data augmentation and consistency loss. 3DIoUMatch benefits from a series of handcrafted threshold and IoU-guided deduplicate module. ProficientTeacher~\cite{yin2022semi} introduces the Spatial-temporal Ensemble module and clustering-based box voting module to improve the recall and precision of pseudo labels. NoiseDet~\cite{Chen_2023_ICCV} views pseudo-labeling as a noisy learning problem and proposes two constraints supervision. 
A-Teacher~\cite{a-teacher} leverages the capability of an offline teacher while jointly updating the entire framework, and achieves significant improvements.
X-Ray Teacher~\cite{gambashidze2024weak} uses multi-frame point cloud to reconstruct the complete shapes of objects to enhance the performance of the teacher model. 
PatchTeacher~\cite{Wu_Peng_Xie_Hou_Lin_Huang_Liu_Cai_Ouyang_2024} leverages partial scene detection to build a super high-resolution teacher for high-quality pseudo labels. 
Unlike previous methods that primarily focus on feature-level consistency or pseudo label quality, our work explicitly explores and leverages the inherent geometric relations within object geometries to guide student learning, offering a complementary perspective for improving SS3D.

\section{Method}
\subsection{Problem definition} 
The objective of semi-supervised 3D object detection is to leverage both labeled and unlabeled data for training detection models. Specifically, the labeled data is represented as $L = \left\{ \left( x^l_i, y^l_i \right) \right\}_{i=1}^{N_l}$, where $x^l_i$ and $y^l_i$ denote the $i$-th point clouds and its corresponding labels, and $N_l$ is the number of labeled samples. In addition to the labeled data, we also have a set of unlabeled data, denoted as $U = \left\{ x^u_i \right\}_{i=1}^{N_u}$, where $N_u$ is the number of unlabeled samples. Typically, $N_u \gg N_l$.
\subsection{Framework Overview}
Our semi-supervised framework consists of two phases. In phase 1, we train a high-performance GeoTeacher. In phase 2, as illustrated in Figure~\ref{fig:pipeline}, GeoTeacher supervises the student with two well-designed modules, Geometric relation supervision and distant-decay voxel-wise data augmentation.

\subsection{Geometric Relation Supervision}
Geometric information not only reflects the surface features of objects but also includes internal structures. Geometric relations help the model understand spatial relationships between different parts of an object, such as relative positions and sizes. This information is crucial for accurate detection and localization, especially with limited labeled data.
In SS3D, geometric information can help the model capture implicit structural characteristics of objects from unlabeled data, thereby improving detection accuracy. Motivated by this, we propose a geometric relation supervision module~(GRS).

To implement geometric relation supervision, we select a set of representative keypoints $\{(x_i, y_i)\}_{i=1}^{K}$ from each object. Considering the common challenges in point cloud data, such as occlusion and sparsity, we select three types of key points: center points, edge midpoints, and corner points based on the bird's-eye-view~(BEV)-projected 2D bounding boxes. 
The center point offer a stable reference for the object’s localization and is less affected by sparse or occlusion. The edge midpoints capture the object’s orientation and spatial extent along both axes, offering some geometric information. The corner points encode fine-grained boundary information, which is essential for delineating the precise spatial extent of the object.
While each individual keypoints encodes local geometric attributes, the relationships among these points capture high-order geometric information that reflects the intrinsic geometry of the object.

To enable effective supervision, we first align the BEV feature maps of the teacher and student models by transforming them into the same coordinate space.
After alignment, we extract geometry-aware features $F^t$ and $F^s$ from both teacher and student models based on the selected keypoints. These features serve as the basis for computing geometric relations, enabling the student to learn geometric knowledge from the teacher in a spatially consistent manner.
We then compute the pairwise geometric relations $M_{\mathrm{rel}}^s$ among keypoints by measuring the cosine similarity between their feature representations. This results in a geometric relation matrix that captures the relative structural dependencies among different parts of the object.
\begin{equation}
    M_{\mathrm{rel}}^s(i,j) = \Phi \left( F^s(x_i, y_i), F^t(x_j, y_j) \right), \quad 1 \leq i, j \leq K,
\end{equation}
where $\Phi$ is the cosine similarity function. The relation matrix $M_{rel}^t$ of the teacher can also be calculated based on $F^t$, and the geometric relation supervision loss is computed as a standard L1 loss:
\begin{equation}
    \mathcal{L}_\delta = \frac{1}{K^2} \sum_{1 \leq i,j \leq K} \left| M_{\mathrm{rel}}^s(i,j) - M_{\mathrm{rel}}^t(i,j) \right|.
\end{equation}
In the semi-supervised training phase, the bounding boxes used for keypoint selection are derived from pseudo labels, which are predicted by the teacher model. However, the quality of these pseudo labels is often suboptimal, especially in challenging scenes with occlusions, sparsity, or overlapping objects.
On the one hand, the inaccurate boxes may lead to the misplacement of the keypoints, resulting in noisy geometric relationships. On the other hand, the feature representations at these keypoints may also be contaminated by background noise, especially when the box includes irrelevant regions. Hence, to mitigate the aforementioned issues, we propose a simple but effective strategy based on confidence-aware weighting. Without bells and whistles, we leverage the classification scores~$\{s_k\}_{k=1}^{n_u}$ predicted by the teacher model to assess the reliability of the pseudo labels, where $s_k$ denotes the score of the $k$-th pseudo-label. Pseudo labels with higher scores are assigned a greater weight in the supervised loss, whereas those with lower scores contribute less. The final loss of geometric relation supervision is computed as follows:
\begin{equation}
    \mathcal{L}_\mathrm{GRS} = \sum_{k=1}^{n_u} s_k~\cdot\mathcal{L}_\delta.
\end{equation}
Furthermore, we introduce an adaptive layer in the student model to ensure it is more complex than the teacher model during training, following~\cite{zhou2023unidistill}.

\subsection{Distant-decay Voxel-wise Data Augmentation}
Although large-scale unlabeled data is available in semi-supervised settings, object geometric diversity is not fully leveraged. This is mainly because pseudo labels tend to focus on objects with regular and complete shapes, while ignoring those with occlusions or rare structural patterns. As a result, the student model lacks exposure to diverse geometric variations, which may hinder its ability to generalize to unseen object structures. 
To alleviate this limitation, we propose a voxel-wise data augmentation strategy~(DVA) that explicitly increases geometric diversity at the object level.

Our data augmentation strategy consists of two operations that construct diverse object representations and effectively enhance the student model’s geometric understanding of unlabeled data, which can be effectively combined with our geometric relation supervision module.
Meanwhile, considering that point clouds of distant objects are inherently sparse, we introduce a distance-decay strategy to reduce the probability of applying data augmentation to distant objects. 
This design contributes to the overall stability of the semi-supervised learning framework by improving detection performance in near-range regions while preserving reliability in distant areas.
The overall logic-based algorithm is illustrated in Algorithm~\ref{alg:voxel_order_dropout}.

Specifically, we first partition the objects based on their bounding boxes into $n_l \times n_w \times n_h$ voxels.
In the voxel-wise sparsify process, we randomly sample the point clouds within the selected voxels, effectively increasing the geometric diversity of the objects. As for voxel-wise order dropout, we randomly remove points from selected voxels, following a spatial sequence in either a clockwise or counterclockwise order. This approach simulates the point cloud distribution of occluded surfaces.
We define that voxels within the same bounding box share the same probability $p$ to participate in the augmentation operations.
To ensure that the point clouds of distant objects remain unaffected, we introduce a distance-decay function to adjust the probability $p$:
\begin{equation}
    p = c \cdot \exp\left( -\frac{\sqrt{i^2 + j^2}}{d_{\mathrm{range}}} \right),
\end{equation}
where $p$ represents the probability of participating in the data augmentation, $c$ is a hyperparameter. $i$ and $j$ are the coordinates of the bounding box center, and $d_{\mathrm{range}}$ denotes the range of the scene. Through our distance-decay voxel-wise augmentation strategies, we construct a richer and more diverse training set that reflects the complexities of a real-world environment, thereby enhancing the detection performance of our models. 
Notably, our method is suitable for both supervised and semi-supervised training settings. For labeled data, we generate the voxels with the ground truth boxes. For unlabeled data, we apply a classification threshold $\tau_\mathrm{aug}$, set higher than the threshold used for pseudo-label selection, to avoid introducing incorrect information.
Furthermore, our data augmentation scheme is plug-and-play, allowing for straightforward integration. 

\subsection{Loss Function}
Finally, we get the total loss of the student model:
\begin{equation}
    \mathcal{L}_\mathrm{total} = \mathcal{L}_\mathrm{base} + \lambda_{1}\cdot\mathcal{L}_\mathrm{GRS},
\end{equation}
where $\mathcal{L}_\mathrm{base}$ is the basic semi-supervised loss, consisting of a regression loss and a classification loss. $\lambda_{1}$ is a hyperparameters used to balance the scale of different losses. 

\begin{algorithm}[t]
\caption{Voxel-wise order dropout}
\label{alg:voxel_order_dropout}
\begin{algorithmic}[1]
\STATE \textbf{Input:} boxes $B$, point cloud $P_\mathrm{i}$, probability $p$, Minimum points to retain $N_\mathrm{p}$, Number of boxes~$N_\mathrm{b}$.
\STATE \textbf{Output:} point cloud $P_\mathrm{o}$

\STATE $voxels \gets \text{get\_voxels}(B)$
\STATE $\textit{v}_\mathrm{c} \gets voxels[..., :3]$
\STATE $\theta_{\mathrm{voxels}} \gets \arctan(\textit{v}_\mathrm{c}[..., 1] / \textit{v}_\mathrm{c}[..., 0])$
\STATE $\theta_{\mathrm{B}} \gets \arctan(B[:, 1] / B[:, 0])$
\STATE \text{Adjust} $\theta_{\mathrm{voxels}}$ \text{relative to} $\theta_\mathrm{B}$, \text{ensuring} \text{that} \text{the} \text{angles} \text{lie} \text{within} $[-\pi, \pi]$
\STATE $mask \gets$ \text{Generate masks based on} $p$
\STATE $indices \gets argsort~(\theta_{\text{voxels}})$

\STATE \text{select a clockwise or counterclockwise order.}

\STATE $will\_drop \gets 0$

\FOR{$i = 1$ \TO $N_\mathrm{b}$}
    \STATE $count \gets \sum(mask[i])$
    \STATE $will\_drop[i, indices[i, :count]] \gets 1$
\ENDFOR

\STATE $\textit{points}_i, \textit{points}_o \gets \text{split\_points}(P_\mathrm{i}, voxels)$

\IF{$\sum(\textit{points}_i) - \sum(\textit{points}_i * will\_drop) > N_\mathrm{p}$}
    \STATE $\textit{points}_i[will\_drop] \gets \emptyset$;
\ENDIF
\STATE $P_\mathrm{o} \gets \text{concatenate}(\textit{points}_o, \textit{points}_i)$
\RETURN $P_\mathrm{o}$
\end{algorithmic}
\end{algorithm}

\section{Experiments}

\begin{table*}[ht]
\setlength{\tabcolsep}{1mm}
\caption{Comparison with other SS3D approaches on ONCE dataset. We combine our GeoTeacher with ProficientTeacher~\cite{Wu_Peng_Xie_Hou_Lin_Huang_Liu_Cai_Ouyang_2024} and ProficientTeacher~\cite{yin2022semi}.}
    \begin{tabular}{cccccccccccccc}
        \toprule[1.5pt]
        \multicolumn{1}{c|}{\multirow{2}{*}{\textbf{Methods}}} &
          \multicolumn{4}{c|}{\textbf{Vehicle AP (\%)}} &
          \multicolumn{4}{c|}{\textbf{Pedestrian AP (\%)}}  &
          \multicolumn{4}{c|}{\textbf{Cyclist AP (\%)}} &
          \multicolumn{1}{c}{\multirow{2}{*}{\textbf{mAP (\%)}}} \\
        \multicolumn{1}{c|}{} &  overall &  0-30m &  30-50m &
          \multicolumn{1}{l|}{50m-inf} &  overall &  0-30m &  30-50m &
          \multicolumn{1}{l|}{50m-inf} &  overall &  0-30m &  30-50m &  
          \multicolumn{1}{l|}{50m-inf} &  \multicolumn{1}{c}{} \\ \hline
        \multicolumn{1}{c|}{Baseline} &  71.19 &  84.04 &  63.02 &
          \multicolumn{1}{l|}{47.25} &  26.44 &  29.33 &  24.05 &
          \multicolumn{1}{l|}{18.05} &  58.04 &  69.96 &  52.43 &
          \multicolumn{1}{l|}{34.61} &  51.89 \\ \hline
        \multicolumn{14}{c}{\textbf{Small} (100K unlabeled Samples)} \\ \hline
        \multicolumn{1}{c|}{Noisy Student} &  73.69 &  84.69 &  67.72 &
          \multicolumn{1}{l|}{53.41} &  28.81 &  33.23 &  23.42 &
          \multicolumn{1}{l|}{16.93} &  54.67 &  65.58 &  50.43 &
          \multicolumn{1}{l|}{32.65} &  52.39\\
        \multicolumn{1}{c|}{SESS} &  73.33 &  84.52 &  66.22 &
          \multicolumn{1}{l|}{52.83} &  27.31 &  31.11 &  23.94 &
          \multicolumn{1}{l|}{19.01} &  59.52 &  71.03 &  53.93 &
          \multicolumn{1}{l|}{36.68} &  53.39\\
        \multicolumn{1}{c|}{3DIoUMatch} &  73.81 &  84.61 &  68.11 &
          \multicolumn{1}{l|}{54.48} &  30.86 &  35.87 &  25.55 &
          \multicolumn{1}{l|}{18.30} &  56.77 &  68.02 &  51.80 &
          \multicolumn{1}{l|}{35.91} &  53.81\\ 
        
        \multicolumn{1}{c|}{NoiseDet} &  75.26 &  86.36 &   67.52 &
          \multicolumn{1}{l|}{55.29} &  37.96 &  42.36 &  32.78 &
          \multicolumn{1}{l|}{23.28} &  60.77 &  72.31 &  55.03 &
          \multicolumn{1}{l|}{38.87} &  58.00\\
        \hline
        \multicolumn{1}{c|}{ProficientTeacher} & 76.07  & 86.78   &70.19   &
          \multicolumn{1}{l|}{56.17} & 35.90  & 39.98  & 31.67  & 
          \multicolumn{1}{l|}{24.37} & 61.19  &73.97   & 55.13  & 
          \multicolumn{1}{l|}{36.98} & 57.72\\
        \multicolumn{1}{c|}{+ GeoTeacher~(Ours)} & \textbf{78.13} & 86.53 & 73.00 &
          \multicolumn{1}{l|}{{60.18}} & \textbf{38.39} & 43.49 & 34.39 &
          \multicolumn{1}{l|}{{21.26}} & \textbf{62.67} & 74.15 & 57.74 &
          \multicolumn{1}{l|}{{40.06}} & \textbf{59.73}\\
        \hline
        
        \multicolumn{1}{c|}{PTPM} & 76.27 & 86.55 & 69.61 &
          \multicolumn{1}{l|}{{56.02}} & 44.29 & 51.95 & 35.86 &
          \multicolumn{1}{l|}{{20.91}} & 61.70 & 75.19 & 54.92 &
          \multicolumn{1}{l|}{{34.57}} & 60.75\\
          
        \multicolumn{1}{c|}{+ GeoTeacher~(Ours)} & \textbf{79.32} & 89.61 & 73.98 &
          \multicolumn{1}{l|}{{58.93}} & \textbf{44.60} & 50.56 & 38.88 &
          \multicolumn{1}{l|}{{27.03}} & \textbf{64.08} & 76.06 & 57.54 &
          \multicolumn{1}{l|}{{40.36}} & \textbf{62.67}\\
          \hline
          
        \multicolumn{14}{c}{\textbf{Medium} (500K unlabeled Samples)} \\ \hline
        \multicolumn{1}{c|}{Noisy Student} &  75.53 &  86.52 &  69.78 &
          \multicolumn{1}{l|}{55.05} &   31.56 &  35.80 &  26.24 &
          \multicolumn{1}{l|}{21.21} &  58.93 &  69.61 &  53.73 &
          \multicolumn{1}{l|}{36.94} &  55.34\\
        \multicolumn{1}{c|}{SESS} &  72.11 &  84.06 &  66.44 &
          \multicolumn{1}{l|}{53.61} &  33.44 &  38.58 &  28.10 &
          \multicolumn{1}{l|}{18.67} &  61.82 &  73.20 &  56.60 &
          \multicolumn{1}{l|}{38.73} &  55.79\\
        \multicolumn{1}{c|}{3DIoUMatch} &  75.69 &  86.46 &  70.22 &
          \multicolumn{1}{l|}{56.06} &  34.14 &  38.84 &  29.19 &
          \multicolumn{1}{l|}{19.62} &  58.93 &  69.08 &  54.16 &
          \multicolumn{1}{l|}{38.87} &  56.25\\
        \multicolumn{1}{c|}{NoiseDet} &  77.14 &   87.21 & 69.94 &
          \multicolumn{1}{l|}{58.44} &  40.45 &  44.91 &  35.71 &
          \multicolumn{1}{l|}{24.11} &  62.59 &  73.04 &  57.98 &
          \multicolumn{1}{l|}{42.67} &  60.06\\

        \hline
        \multicolumn{1}{c|}{ProficientTeacher} &  78.07 &  87.43 &72.50   & 
          \multicolumn{1}{l|}{59.51} &38.38  & 42.45  & 34.62   & 
          \multicolumn{1}{l|}{25.58} &63.23  & 74.70 & 58.19  & 
          \multicolumn{1}{l|}{40.73} &59.89\\
        \multicolumn{1}{c|}{+ GeoTeacher~(Ours)} &\textbf{78.94} &87.34 &73.41 &
          \multicolumn{1}{l|}{{60.70}} &\textbf{42.00} &48.11 &35.55 &
          \multicolumn{1}{l|}{{21.77}} &\textbf{63.94} &73.93 &58.71 &
          \multicolumn{1}{l|}{{43.38}} &\textbf{61.62} \\
        \hline
        \multicolumn{1}{c|}{PTPM} &76.66 &86.75 &71.30 &
          \multicolumn{1}{l|}{{56.87}} &45.87 &54.98 &37.35 &
          \multicolumn{1}{l|}{{20.89}} &61.88 &74.08 &56.52 &
          \multicolumn{1}{l|}{{33.30}} &61.47\\ 
        \multicolumn{1}{c|}{+ GeoTeacher~(Ours)} &\textbf{79.51} &89.77 &74.11 &
          \multicolumn{1}{l|}{{59.12}} &\textbf{48.33} &54.91 &41.19 &
          \multicolumn{1}{l|}{{25.83}} &\textbf{64.76} &76.57 &58.57 &
          \multicolumn{1}{l|}{{40.64}} &\textbf{64.20} \\
          \hline
          
        \multicolumn{14}{c}{\textbf{Large} (1M unlabeled Samples)} \\ \hline
        \multicolumn{1}{c|}{Noisy Student} &  75.99 &  86.67 &  70.48 &
          \multicolumn{1}{l|}{55.60} &  33.31 &  37.81 &  28.19 &
          \multicolumn{1}{l|}{21.39} &  59.81 &  70.01 &  55.13 &
          \multicolumn{1}{l|}{38.33} &  56.37\\
        \multicolumn{1}{c|}{SESS} &  75.95 &   86.83  &  70.45 &
          \multicolumn{1}{l|}{55.76} &  34.43 &  40.00 &   27.92 &
          \multicolumn{1}{l|}{19.20} &  63.58 &  74.85 &  58.88 &
          \multicolumn{1}{l|}{39.51} &  57.99\\
        \multicolumn{1}{c|}{3DIoUMatch} &  75.81 &  86.11 &  71.82 &
          \multicolumn{1}{l|}{57.84} &  35.70 &  40.68 &  30.34 &
          \multicolumn{1}{l|}{21.15} &  59.69 &  70.69 &  54.92 &
          \multicolumn{1}{l|}{39.08} &  57.07\\
        \multicolumn{1}{c|}{NoiseDet} &   78.02 &  87.00 &  72.55 &
          \multicolumn{1}{l|}{59.49} &  42.89 &  46.52 &  38.21 &
          \multicolumn{1}{l|}{ 26.60} &  62.74 &  73.19 &  58.03 &
          \multicolumn{1}{l|}{42.88} &  61.16\\
        \hline
        \multicolumn{1}{c|}{ProficientTeacher} & 78.12  &87.22   &72.74   &
          \multicolumn{1}{l|}{59.58} & 41.95  &48.09   &35.13   &
          \multicolumn{1}{l|}{26.01} & 64.12 &75.85   &58.04   &
          \multicolumn{1}{l|}{41.45} & 61.40  \\
        \multicolumn{1}{c|}{+ GeoTeacher~(Ours)} &\textbf{80.86} &89.15 &75.77 &
          \multicolumn{1}{l|}{{63.24}} &\textbf{43.03} &49.93 &35.66 &
          \multicolumn{1}{l|}{{22.70}} &\textbf{65.59} &76.40 &61.18 &
          \multicolumn{1}{l|}{{44.30}} &\textbf{63.16}\\
        \hline
        \multicolumn{1}{c|}{PTPM} &76.46 &86.35 &71.31 &
          \multicolumn{1}{l|}{{57.08}} &45.72 &55.00 &36.81 &
          \multicolumn{1}{l|}{{20.25}} &65.87 &77.41 &59.85 &
          \multicolumn{1}{l|}{{42.39}} &62.68\\

        \multicolumn{1}{c|}{+ GeoTeacher~(Ours)} &\textbf{79.72} &89.86 &74.53 &
          \multicolumn{1}{l|}{{59.22}} &\textbf{51.36} &61.17 &42.69 &
          \multicolumn{1}{l|}{{25.06}} &\textbf{66.03} &77.49 &60.04 &
          \multicolumn{1}{l|}{{40.53}} &\textbf{65.70}\\
        \bottomrule[1.5pt]
    \end{tabular}
\centering

\label{tab:once_second}
\end{table*}
\subsection{Datasets}
\noindent\textbf{ONCE Dataset.}  ONCE~\cite{NEURIPS67c6a1e7} is a large-scale autonomous driving dataset with 1 million LiDAR point cloud samples. Only 15,000 samples are annotated, which have been divided into training, validation, and testing splits with 5k, 3k, and 8k samples, respectively. According to the official SSL setting, 5k labeled samples and all unlabeled samples have been divided into 3 subsets: Small, Medium, and Large. The small set contains 100k samples~(70 sequences), the medium set contains 500k samples~(321 sequences), and the large set contains about 1M samples~(560 sequences). In addition, three different perception ranges, ``0-30m'', ``30-50m'', and ``50m-inf'', are specified to evaluate the performance of 3D detectors.

\noindent\textbf{Waymo Open Dataset.}  Waymo~\cite{Sun_2020_CVPR} is a large-scale LiDAR point cloud dataset, which contains 798 sequences for training and 202 sequences for validation. Following ProficientTeacher~\cite{yin2022semi}, we divide the 798 training sequences equally into labeled split $\mathcal{P^L}$ and unlabeled split $\mathcal{P^U}$, with each containing 399 sequences. $5\%$ and $20\%$ sequences are randomly sampled from $\mathcal{P^L}$, leading to the ratio of labeled data and unlabeled data $\mathcal{P^L} : \mathcal{P^U}$ as 1:20 and 1:5, respectively. 

\subsection{Implementation details}
Since our approach is generic and can be seamlessly combine with other SS3D methods, we select two representative SS3D methods, ProficientTeacher~\cite{yin2022semi} and PTPM~\cite{Wu_Peng_Xie_Hou_Lin_Huang_Liu_Cai_Ouyang_2024}, to validate its effectiveness. In phase 1, we follow the same training strategy as theirs to train the teacher model.
In phase 2, for the training of the student, we sampling 1 labeled and 4 unlabeled samples for each mini-batch to avoid training collapse. The student is trained for 50, 75, and 100 epochs for small, medium, and large settings on the ONCE dataset~\cite{NEURIPS67c6a1e7}. For DVA, we empirically divide each box into $4 \times 2 \times 1$ voxels, set $c$ in DVA and $\tau_\mathrm{aug}$ to 0.05 and 0.7. 
While $\lambda_u$ and $\lambda_1$ are set to 1.0, 2.0, respectively. 
All models are trained on a machine with 4 NVIDIA H100 GPUs.

\subsection{Main results}
\textbf{ONCE dataset.} 
We combine our GeoTeacher with ProficientTeacher~\cite{yin2022semi} and PTPM~\cite{Wu_Peng_Xie_Hou_Lin_Huang_Liu_Cai_Ouyang_2024} and evaluate them on ONCE dataset~\cite{NEURIPS67c6a1e7}.
As shown in Table~\ref{tab:once_second}, our method consistently brings significant improvements under different proportions of unlabeled data. For example, under the small protocol, it improves ProficientTeacher by +2.01 mAP and PTPM by +1.92 mAP. 
Our GeoTeacher also achieves 63.16 mAP and 65.70 mAP under the large protocol, outperforming ProficientTeacher and PTPM by +1.76 and +3.02 mAP, respectively.
It is worth noting that when combined with PTPM, our method achieves 62.67 mAP under the small protocol~(100K unlabeled data), which is comparable to the performance of PTPM under the large protocol~(1M unlabeled data).
This indicates that our geometry-guided method can effectively enhance the utilization of unlabeled data.
\begin{table}[t]
\setlength{\tabcolsep}{5mm}
    \centering
    \caption{Generalizability on different detectors. We combine our GeoTeacher with ProficientTeacher~\cite{Wu_Peng_Xie_Hou_Lin_Huang_Liu_Cai_Ouyang_2024} and ProficientTeacher~\cite{yin2022semi}.}
        \begin{tabular}{c|c|c}
        \toprule[1.5pt]
        \multicolumn{1}{c|}{\multirow{1}{*}{Detector}} &
        \multicolumn{1}{c|}{\multirow{1}{*}{Method}} &        
        \multicolumn{1}{c}{mAP~$\%$} \\
        \hline
        \multirow{4}{*}{PV-RCNN} & ProficientTeacher & 61.27 \\ 
        & + GeoTeacher~(Ours) & \textbf{62.83} \\ 
        \cline{2-3}
        {}& PTPM & 62.22 \\ 
        & + GeoTeacher~(Ours) & \textbf{63.48} \\ 
        \hline
        \multirow{4}{*}{CenterPoint} & ProficientTeacher & 66.84 \\ 
        & + GeoTeacher~(Ours) & \textbf{68.23} \\
        \cline{2-3}
        {}& PTPM & 68.40 \\ 
        & + GeoTeacher~(Ours) & \textbf{69.55} \\ 
        \bottomrule[1.5pt]
        \end{tabular}
    
    \label{tab:Generalization}
\end{table}

\begin{table*}[t]
\centering
\caption{Comparison with other SS3D approaches on Waymo dataset. We use the same data split as PTPM~\cite{Wu_Peng_Xie_Hou_Lin_Huang_Liu_Cai_Ouyang_2024}.}
\setlength{\tabcolsep}{4mm}
        \begin{tabular}{c|c|cccc}
        \toprule[1.5pt]
         \multirow{2}{*}{Label Amounts}  & \multirow{2}{*}{Method} &\multicolumn{4}{c}{3D AP/APH @0.7 (LEVEL 2)}\\
         &  & Overall & Vehicle & Pedestrian & Cyclist \\ \hline
        \multirow{3}{*}{~~~5\% ($\sim$ 4k Labels)}  
        &{SECOND~\cite{yan2018second}}  & 45.78/40.40    & 50.03/49.52   & 45.77/34.98 & 41.53/36.69 \\
        &{HASS~\cite{zeng2024hardness}}          & 52.64/48.27    & 53.36/52.73   & 50.15/39.63 & 54.42/52.46 \\ 
        \cline{2-6}
        &{ProficientTeacher~\cite{yin2022semi}}  & 51.10/45.75    & 53.04/52.54   & 50.33/38.67 & 49.92/\textbf{46.03} \\
        \multirow{2}{*}{~~~$\mathcal{P}^{L}:\mathcal{P}^{U}=1:20$}
        &+ GeoTeacher(Ours)   &\textbf{52.38/47.07}  &\textbf{55.28/54.69}  & \textbf{51.70/41.44}    & \textbf{50.18}/45.09   \\
        \cline{2-6}
        &{PTPM~\cite{Wu_Peng_Xie_Hou_Lin_Huang_Liu_Cai_Ouyang_2024}}  & 54.53/51.07    & 56.98/56.48   & 52.64/44.19 & 53.96/52.55 \\
        &+ GeoTeacher(Ours)   &\textbf{55.45/51.88}  &\textbf{57.84/57.53}   & \textbf{53.65/45.01}    & \textbf{54.88/53.10}   \\
        \hline
        \multirow{3}{*}{~~~20\% ($\sim$ 16k Labels)}  
        &{SECOND~\cite{yan2018second}}  & 53.09/49.11   & 57.40/56.81   & 51.54/41.91   & 50.33/48.62   \\ 
        &{HASS~\cite{zeng2024hardness}}          & 59.09/54.22   & 59.19/58.38   & 57.75/46.87   & 60.32/57.41 \\ 
        \cline{2-6}
        &{ProficientTeacher~\cite{yin2022semi}}  & 58.59/54.16   & 59.97/59.36   & 57.88/46.97   & 57.93/56.15   \\
        \multirow{2.}{*}{~~~$\mathcal{P}^{L}:\mathcal{P}^{U}=1:5$}
        &+ GeoTeacher~(Ours)     & \textbf{59.38/55.60}   & \textbf{61.54/61.04}	& \textbf{58.51/49.39}   &\textbf{58.11/56.37} \\
        \cline{2-6}
        &{PTPM~\cite{Wu_Peng_Xie_Hou_Lin_Huang_Liu_Cai_Ouyang_2024}}  & 60.49/57.02   & 61.62/61.16   & 59.78/51.17   & 60.07/58.74   \\
        &+ GeoTeacher~(Ours)     & \textbf{61.05/57.40}   & \textbf{62.30/61.61}	& \textbf{60.29/51.49}   &\textbf{60.56/59.10} \\
        \hline
        \multirow{4.}{*}{~~~100\% ($\sim$ 80k Labels)} 
        &{SECOND~\cite{yan2018second}}    & 48.80/43.35    & 51.87/51.27   & 48.28/36.56 & 46.26/42.21 \\
        \cline{2-6}
        &{ProficientTeacher~\cite{yin2022semi}}  & 62.96/59.14   & 63.56/63.06   & 62.34/53.19   & 62.97/61.18 \\
        &+ GeoTeacher~(Ours)  &\textbf{65.21/61.55}  & \textbf{66.66/66.19}	& \textbf{65.22/56.45}	& \textbf{63.77/62.03} \\
        \cline{2-6}
        \multirow{2.}{*}{~~~$\mathcal{P}^{L}:\mathcal{P}^{U}=1:5$}
        &{PTPM~\cite{Wu_Peng_Xie_Hou_Lin_Huang_Liu_Cai_Ouyang_2024}}  & 65.73/62.13   & 67.12/66.67   & 65.85/57.11   & 64.23/62.60   \\
        &+ GeoTeacher~(Ours)  &\textbf{66.40/62.52}  & \textbf{67.80/66.97}	& \textbf{66.59/57.46}	& \textbf{64.80/63.14} \\
        \cline{2-6}
        {}& {~~Oracle Model }& 62.43/58.87     & 65.43/64.91   & 61.82/52.93  & 60.03/58.76\\
        \bottomrule[1.5pt]
        \end{tabular}

\label{tab:waymo_second}
\end{table*}

\noindent\textbf{Generalization on other detectors.} 
To better assess the generalization capability fo our method, we combined GRS and DVA with ProficientTeacher~\cite{yin2022semi} and PTPM~\cite{Wu_Peng_Xie_Hou_Lin_Huang_Liu_Cai_Ouyang_2024}, then conducted experiments on two additional detectors: PV-RCNN~\cite{Shi_2020_CVPR}, a classic two-stage detector, and CenterPoint~\cite{Yin_2021_CVPR}, a widely adopted single-stage anchor-free detector. Experiments are conducted on the ONCE dataset using the small protocol. As shown in Table~\ref{tab:Generalization}, our method yields significant improvements across different detectors. Using PV-RCNN, our method achieves improvements of 1.56 mAP and 1.26 mAP, respectively.
Despite CenterPoint being a stronger detector, our method still delivers a remarkable improvement of +1.39 mAP and +1.15 mAP, respectively. These results demonstrate the robust generalization ability of our approach.

\noindent\textbf{Waymo Open dataset.} 
We further evaluate our method with existing methods on the Waymo Open dataset~\cite{Sun_2020_CVPR} with ProficientTeacher~\cite{yin2022semi} and PTPM~\cite{Wu_Peng_Xie_Hou_Lin_Huang_Liu_Cai_Ouyang_2024}. The experimental results in Table~\ref{tab:waymo_second} show that our GeoTeacher consistently improve two famous methods under different labeling ratios. 
In the widely used 5\% label regime, our approach outperforms PTPM by +0.92 AP and +0.81 APH and ProficientTeacher by +1.28 AP and +1.32 APH, and this advantage remains evident under the 20\% labeled setting, further validating the robustness and scalability of our framework.  Moreover, we also run an oracle model that trains the detector with the full 798 sequences, while our model with only half labels outperforms it.
These demonstrate that by learning geometric information, the student model can achieve superior detection performance even on a complex dataset like Waymo. 
To further evaluate our method, we use PV-RCNN~\cite{Shi_2020_CVPR} as baseline detector and compare with HSSDA~\cite{liu2023hierarchical} and PTPM. As shown in Table~\ref{tab:waymo_pvrcnn}, our approach also delivers a notable improvement over previous methods.

\begin{table}[t!]
\caption{Performance comparison on the Waymo Open Dataset with 202 validation sequences for the 3D object detection with PV-RCNN~\cite{shi2023pv} as the detector.}
\centering
        \setlength\tabcolsep{1.mm}{\begin{tabular}{c|cccc}
        \toprule
         \multirow{2}{*}{\begin{tabular}[c]{@{}c@{}}1\% Data \\ ($\sim$ 1.4k scenes)\end{tabular}} &\multicolumn{3}{c}{3D AP/APH @0.7 (LEVEL 2)} \\
           & Vehicle & Pedestrian & Cyclist \\ \hline
        PV-RCNN  & 44.7/ 42.2 &30.7 / 15.3 &2.8 / 1.6 \\ 
        DetMatch  &48.1 / 47.2 &35.8 / 17.1 & - / - \\ 
        HSSDA  &49.7 / 47.3 &33.5 / 17.5 & 27.9/ 20.0 \\ 
        \cline{1-4}
        PTPM&53.7 / 52.2 &36.3 / 18.8 &35.7 / 34.3 \\ 
        + GeoTeacher~(Ours) &\textbf{55.3} / \textbf{53.6} &\textbf{38.1} / \textbf{19.7} &\textbf{37.2} / \textbf{35.8} \\ 
        \bottomrule
        \end{tabular}
        }
        \vspace{-5pt}
\label{tab:waymo_pvrcnn}
\end{table}

\begin{table}[t]
    \centering
    \caption{Ablation study on each component of GeoTeacher.}
    \setlength{\tabcolsep}{1.5mm}
        \begin{tabular}{cc|cccc}
        \toprule[1.5pt]
        \multirow{2}{*}{DVA} & \multirow{2}{*}{GRS} & \multicolumn{4}{c}{mAP~($\%$)} \\
        & & Overall & Vehicle & Pedestrian & Cyclist \\
        \midrule
        & &60.06 &76.45 & 41.95 & 61.78   \\ 
        \ding{51} & & 61.45 & 77.35 & 43.11 & 63.91 \\
        & \ding{51} & 61.68 & 78.14 & 43.62 & 63.29 \\ 
        \ding{51} & \ding{51} & \textbf{62.67} & \textbf{79.32} & \textbf{44.60} & \textbf{64.08} \\
        \bottomrule[1.5pt]
        \end{tabular}
    \label{tab:student_ablation}
\end{table}
\subsection{Ablation Study}
To better understand how the proposed approach works, we conduct a series of ablation studies on the ONCE small protocol training set, with evaluations on the validation set.

\noindent\textbf{The effect of each component.}
We conduct ablation studies to evaluate the individual and combined effects of our proposed Distant-decay voxel-wise data augmentation and Geometric Relation Supervision. 
We combine PTPM with our method, and the result obtained by removing GRS and DVA is used as the baseline.
As shown in Table~\ref{tab:student_ablation}, each module independently contributes to performance improvement, and their combination yields the highest mAP.
Specifically, applying DVA alone boosts the overall mAP by +1.39, indicating its effectiveness in enriching the geometric diversity of objects from unlabeled data. GRS alone also brings a comparable gain of +1.62 mAP, demonstrating the benefit of explicitly modeling object-internal geometric relations. When both components are combined, our model achieves the best result, showing that the two components are complementary. Together, they enhance the student model’s ability to perceive and understand geometric cues in the semi-supervised setting. These results verify that both data-level and supervision-level geometry modeling are crucial to the success of GeoTeacher.
\begin{table}[ht]
    \centering
    \caption{Comparison of our DVA and other approaches.}
    \setlength{\tabcolsep}{1mm}
        \begin{tabular}{c cccc}
         \toprule[1.5pt]
         \multirow{2}{*}{Method}& \multicolumn{4}{c}{mAP~($\%$)} \\
         & Overall & 0-30m & 30-50m & 50m-inf \\
         \midrule
        Baseline &53.34 &62.44 & 47.52 &33.38 \\
        SE-SSD   & 54.79 &64.31 & 50.18 & 34.97 \\ 
        TED  &55.22 &64.44 &50.27 &35.12 \\
        HINTED &54.84 &64.52 &49.44 &35.51 \\
        DVA~(Ours)& \textbf{55.45} &\textbf{64.79} &\textbf{50.55} & \textbf{36.19}\\
        \bottomrule[1.5pt]
        \end{tabular}
    
    \label{tab:dva_abl}
\end{table}

\noindent\textbf{Comparison with DVA and other approaches.}
To validate the superiority of our DVA, we compared it with other related methods and presented the results at different distances. We adopt three well-designed data augmentation strategies for comparisons: 1)~SE-SSD~\cite{zheng2021se} divides object’s point cloud into six pyramidal-shape subsets and then undergo random sparsify, random dropout, and random swapping; 2)~TED~\cite{wu2023transformation} augments objects by translating the point clouds along the depth axis and down-sampling the whole object point clouds; 3)~HITED~\cite{xia2024hinted} down-samples nearby point clouds to simulate distant scenes.
We use SECOND~\cite{yan2018second}, trained solely on labeled data, as our baseline and incorporate the aforementioned augmentation modules into it.
The results in Table~\ref{tab:dva_abl} show that each approach improves the baseline, while our DVA achieves superior performance. Our DVA brings +2.11 mAP overall and gains significant improvement, which indicates that increasing geometric diversity through our augmentation strategy remains beneficial even in the presence of labeled data. Moreover, our method achieves superior performance in the ``50–inf'' range, which can be attributed to the proposed distant-decay mechanism.
\begin{table}[ht]
    \centering
    \caption{Comparison of our GRS and other approaches.}
    \setlength{\tabcolsep}{1mm}
        \begin{tabular}{c cccc}
         \toprule[1.5pt]
         \multirow{2}{*}{Method}& \multicolumn{4}{c}{mAP~($\%$)} \\
         &  Overall & Vehicle & Pedestrian & Cyclist \\
         \midrule
            Baseline &60.06 &76.45 & 41.95 & 61.78 \\
            SOOD & 60.31 &76.97 & 41.94 &62.01 \\ 
            NoiseDet & 61.10&78.12 & 42.63 &62.55 \\
            BoxMask  &60.84 & 78.90 & 42.39 & 61.23 \\
            GRS~(Ours)&\textbf{61.68} & \textbf{78.14} & \textbf{43.62} & \textbf{63.29}  \\
        \bottomrule[1.5pt]
        \end{tabular}
    
    \label{tab:rws_box_loss}
\end{table}

\noindent\textbf{Comparison with GRS and other approaches.} 
Unlike common semi-supervised approaches that rely on feature distillation, our proposed GeoTeacher guides the student model through geometric relation modeling. To demonstrate the superiority of our approach, we compare it with several representative feature-level distillation methods: SOOD~\cite{hua2023sood}, NoisDet~\cite{Chen_2023_ICCV} and the commonly used box-level masking method.
Although our GRS does not directly enforce consistency at the feature level, it transfers high-order geometric knowledge between the teacher and student models through relational modeling among keypoints. Unlike prior methods that treat features in isolation, our approach captures the internal geometric relation within objects.
As shown in Table~\ref{tab:rws_box_loss}, the aforementioned methods have achieved significant improvements compared to the baseline, our GRS attains the highest performance, surpassing the other methods by at least 0.58 mAP overall. These results demonstrate that geometric relations can be more effective than enforcing low-level feature similarity, as it provides richer inductive biases that are crucial for understanding the spatial organization of objects, particularly in sparse or occluded point cloud scenarios.

\begin{table}[h]
    \centering
    \caption{Ablations of grid-set of DVA.}
    \setlength{\tabcolsep}{1.00mm}
        \begin{tabular}{c cccc}
             \toprule[1.5pt]
             Grid-Set& \multicolumn{4}{c}{mAP~($\%$)} \\
             ($n_l \times n_w \times n_h$) &  Overall & Vehicle& Pedestrian & Cyclist \\
             \midrule
            $ 3 \times 3 \times 2$  & 55.08 &73.17 & 32.33 &\textbf{59.72} \\  
            $ 3 \times 2 \times 1$  & 55.27 & 74.67 & \textbf{32.40} & 58.73  \\ 
            $ 4 \times 2 \times 1$  & \textbf{55.45} &\textbf{74.71} &32.33 & 59.31 \\
            $ 4 \times 2 \times 2$  & 54.79 &73.68 &31.68 &59.01 \\
            \bottomrule[1.5pt]
        \end{tabular}
    \label{tab:dva_voxelsize}
    \vspace{-15pt}
\end{table}

\begin{table}[h]
\caption{Ablations of key points selection of GRS.}
    \centering
        \begin{tabular}{cccc}
         \toprule[1.5pt]
            Center& Corner &Edge & Overall  \\
            points & points & midpoints & mAP~(\%)\\
             \midrule
            \checkmark& \checkmark& & 62.09  \\
            & \checkmark& \checkmark &61.84\\ 
            \checkmark& & \checkmark& 62.14  \\ 
            \checkmark& \checkmark& \checkmark& \textbf{62.67} \\ 
        \bottomrule[1.5pt]
        \end{tabular}
    
    \label{tab:keypoint_ablation}
\end{table}
\noindent\textbf{The effect of hyperparameters.}
We now investigate the effect of different hyperparameters in our framework.
We first ablate the grid-set on our data augmentation strategy. As shown in Table~\ref{tab:dva_voxelsize}, all grid-set configurations lead to performance improvements compared to the baseline presented in Table~\ref{tab:dva_abl}. Notably, the best performance is achieved with the $4\times2\times1$ grid-set configuration. Subsequently, we validate the impact of keypoints selection in GRS. The results are shown in Table~\ref{tab:keypoint_ablation}. All combinations of keypoints lead to performance gains, and the best results are achieved when center points, edge midpoints, and corner points are used together. This demonstrates that the selected keypoints provide complementary geometric information, and jointly modeling their relations yields more effective geometric relation supervision.
We also conducted hyperparameter experiments for the loss weight $\lambda_1$ and the pseudo-label confidence threshold $c$. As shown in Table~\ref{tab:ablation_supp_GRS} and Table~\ref{tab:ablation_supp_threshold} determining that the model's performance is optimal when they are set to 2.0 and 0.3, respectively.
\begin{figure}[t]
    \centering
    \includegraphics[width=1.0\linewidth]{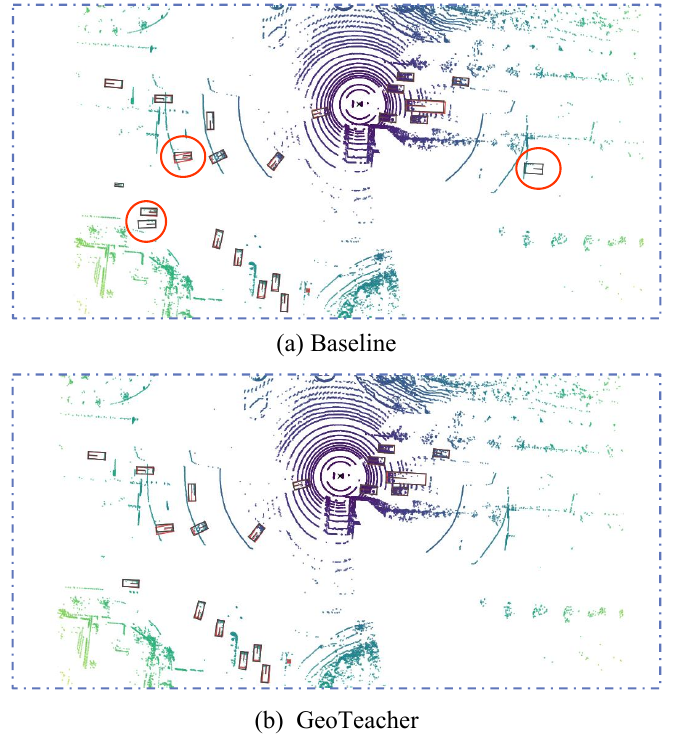}
    \caption{
    Some visualization examples from the ONCE dataset~\cite{NEURIPS67c6a1e7} are presented. From top to down, each row displays the baseline results and the predictions of our GeoTeacher. Red rectangles denote ground truth, black rectangles represent predictions, and red dashed circles highlight the improvements achieved by our method.
    }
    \label{fig:vis}
\end{figure}
\begin{table}[t]
    \centering
    \caption{The effect of loss weight $\lambda_1$.}
    \begin{tabular}{c|c|c|c|c}
        \toprule[1.5pt]
        $\lambda_1$ & 0.5 & 1.0 & 2.0 & 3.0 \\
        \midrule
        Overall~mAP~($\%$) & 60.97 & 61.22 & \textbf{61.68} & 60.54 \\
        \bottomrule[1.5pt]
    \end{tabular}
    \label{tab:ablation_supp_GRS}
\end{table}

\begin{table}[h!]
    \centering
    \caption{The effect of classification threshold $c$.}
    \begin{tabular}{c|c|c|c|c}
        \toprule[1.5pt]
        $c$ & 0.2 & 0.3 & 0.4 & 0.5 \\
        \midrule
        Overall~mAP~($\%$) & 54.75 & \textbf{55.53} & 55.21 & 54.90 \\
        \bottomrule[1.5pt]
    \end{tabular}
    \label{tab:ablation_supp_threshold}
\end{table}

\subsection{Qualitative visualization results}
In this section, we present qualitative results to demonstrate the effectiveness of GeoTeacher. As shown in Figure~\ref{fig:vis}, we compare our method with the baseline model. The visualization reveals that our approach significantly reduces false positives and better adjusts object orientations, highlighting the benefits of the proposed method.

\section{Conclusion}
In this paper, we present \textit{GeoTeacher}, a novel method for semi-supervised 3D object detection. 
We analyze the importance of object's geometric information for SS3D, particularly in the context of limited labeled data and large-scale unlabeled data. To this end, we propose a Geometric Relation Supervision that guides the student model to learn the intrinsic geometric relation of objects, leading to significant performance improvements. In addition, to further enhance geometric diversity, we introduce a distant-decay voxel-wise augmentation strategy that perturbs the geometric relationships of nearby objects more aggressively while preserving the detectability of distant ones.
Extensive experiments on the ONCE and Waymo datasets demonstrate the effectiveness and generalization of our approach. 

\section{Acknowledgements} 
This work was supported in part by New Generation Artificial Intelligence-National Science and Technology Major
Project (2025ZD0123004), Ningbo grant (2025Z038) and National Natural Science Foundation of China (Grant No. 62376060).

{\small
\bibliographystyle{IEEEtran}
\bibliography{IEEEabrv}
}



\addtolength{\textheight}{-12cm}   


\end{document}